\documentclass[conference]{IEEEtran}
\def\IEEEtitletopspace{18pt}

\usepackage[top=54pt, left=54pt, right=54pt, bottom=54pt]{geometry}
\IEEEoverridecommandlockouts
\usepackage{amsmath,amssymb,amsfonts}
\usepackage{algorithmic}
\usepackage{multicol}
\usepackage{graphicx}
\usepackage{hyperref}
\usepackage[numbers]{natbib}
\usepackage{textcomp}
\usepackage{xcolor}
\hypersetup{bookmarksopen,bookmarksnumbered,
pdfpagemode=UseOutlines,
colorlinks=true,
urlcolor=black,
linkcolor=red,
anchorcolor=red,
citecolor=red,
filecolor=red,
menucolor=red
}
\usepackage{cleveref}
\usepackage{booktabs}
\usepackage{tikz}
\usepackage{siunitx}
\usepackage[font=footnotesize]{caption}
\usepackage{subcaption}
\usepackage{nicefrac}
\usepackage{soul}
\usepackage{multirow}

\def\BibTeX{{\rm B\kern-.05em{\sc i\kern-.025em b}\kern-.08em
    T\kern-.1667em\lower.7ex\hbox{E}\kern-.125emX}}
\begin{document}

\title{\LARGE \bf Scalable Learning of Segment-Level Traffic Congestion Functions}

\makeatletter
\newcommand{\linebreakand}{%
  \end{@IEEEauthorhalign}
  \hfill\mbox{}\par
  \mbox{}\hfill\begin{@IEEEauthorhalign}
}
\makeatother

\author{Shushman Choudhury$^{1,2}$, Abdul Rahman Kreidieh$^{2}$, Iveel Tsogsuren$^{2}$,\\ Neha Arora$^{2}$, Carolina Osorio$^{2,3}$, and Alexandre M. Bayen$^{4,5}$
\thanks{$^{1}$Corresponding author, shushmac@google.com}%
\thanks{$^{2}$Google Research}%
\thanks{$^{3}$HEC Montr\'eal}%
\thanks{$^{4}$University of California, Berkeley}%
\thanks{$^{5}$Done while at Google Research}
}

\maketitle

\begin{abstract}
We propose and study a data-driven framework for identifying traffic congestion functions (numerical relationships between observations of traffic variables) at global scale and segment-level granularity. In contrast to methods that estimate a separate set of parameters for each roadway, ours learns a single black-box function over all roadways in a metropolitan area. First, we pool traffic data from all segments into one dataset, combining static attributes with dynamic time-dependent features. Second, we train a feed-forward neural network on this dataset, which we can then use on any segment in the area. 
We evaluate how well our framework identifies congestion functions on observed segments and how it generalizes to unobserved segments and predicts segment attributes on a large dataset covering multiple cities worldwide. For identification error on observed segments, our single data-driven congestion function compares favorably to segment-specific model-based functions on highway roads, but has room to improve on arterial roads. For generalization, our approach shows strong performance across cities and road types: both on unobserved segments in the same city and on zero-shot transfer learning between cities. Finally, for predicting segment attributes, we find that our approach can approximate critical densities for individual segments using their static properties.
\end{abstract}


\section{Introduction}
\label{sec:intro}

We consider the problem of identifying segment-level traffic congestion functions, which are numerical relationships between observed values of the key traffic variables: flow, speed, and density. For any road segment, given potentially partial observations of one variable in a time interval, we want to estimate another (a system identification problem~\cite{ljung1998system}). In particular, our approach learns from data without assumptions on traffic physics and seeks a global solution that can scale to the whole world.

Traffic is a critical component of urban transportation and affects safety, sustainability, and quality of life. Congestion functions help understand the effect of varying observed vehicle volume on travel time. Good estimates of these effects, especially if scaled across urban areas, would enable and improve many critical transportation applications, e.g., real-time navigation systems could better adapt to changing network-wide congestion patterns and transportation policy-makers could better predict the impact of major interventions~\cite{deng2021systematic}.

Identifying segment-level relationships between traffic variables at massive scale is difficult primarily due to the sparsity of data. The transportation community typically focuses on the similar but distinct problem of estimating fundamental diagrams, which relates observed \emph{total} flow, speed and density with assumed functional forms~\cite{daganzo1997fundamentals}. The vast majority of such fundamental diagram approaches use noisy data from fixed sensors that work only on specific roadways. ~\citet{bramich2022fitting} benchmarked several functional forms on loop detector data from multiple cities, but each dataset only covered a few dozen roads per city. Such approaches could struggle to scale to all segments in a large area; the best functional form may vary and several segments may have insufficient datapoints for an individual fit.

Floating vehicle data from GPS tracks can have much wider spatial coverage than static infrastructure. Its key limitation is varying vehicle penetration rates, which makes it difficult, if not impossible, to estimate total flow~\cite{treiber2013trajectory}. However, scalability is paramount to us, so we accept and proceed with the limitation of only observing \emph{partial} flow. We use the term congestion function (distinct from fundamental diagram) to refer to the relationship between observed partial flow, speed, and density. Until now, the transportation research community has lacked large-scale, i.e., city-wide, empirical evaluations of segment-level congestion functions, especially the downstream predictive utility on unobserved data. This article will address that gap in the literature.

\newcommand\neuralnetwork[3]{
    \def\x{#1}
    \def\y{#2}
    \def\radius{#3}

    \draw[->,thick] (\x,\y) -- (\x+2*\radius,\y);
    \draw[->,thick] (\x,\y+3*\radius) -- (\x+2*\radius,\y+3*\radius);
    \draw[->,thick] (\x,\y-3*\radius) -- (\x+2*\radius,\y-3*\radius);
    \filldraw[color=black, fill=gray!25, thick](\x+3*\radius,\y) circle (\radius);
    \filldraw[color=black, fill=gray!25, thick](\x+3*\radius,\y+3*\radius) circle (\radius);
    \filldraw[color=black, fill=gray!25, thick](\x+3*\radius,\y-3*\radius) circle (\radius);

    \filldraw[color=black, fill=gray!25, thick](\x+7*\radius,\y-1.5*\radius) circle (\radius);
    \filldraw[color=black, fill=gray!25, thick](\x+7*\radius,\y+1.5*\radius) circle (\radius);
    \filldraw[color=black, fill=gray!25, thick](\x+7*\radius,\y+4.5*\radius) circle (\radius);
    \filldraw[color=black, fill=gray!25, thick](\x+7*\radius,\y-4.5*\radius) circle (\radius);
    \foreach \i in {0,...,2}{
        \foreach \j in {0,...,3}{
            \draw[]
            (\x+4*\radius, \y-3*\radius+3*\i*\radius) --
            (\x+6*\radius, \y-4.5*\radius+3*\j*\radius);
        }
    }

    \filldraw[color=black, fill=gray!25, thick](\x+11*\radius,\y-1.5*\radius) circle (\radius);
    \filldraw[color=black, fill=gray!25, thick](\x+11*\radius,\y+1.5*\radius) circle (\radius);
    \filldraw[color=black, fill=gray!25, thick](\x+11*\radius,\y+4.5*\radius) circle (\radius);
    \filldraw[color=black, fill=gray!25, thick](\x+11*\radius,\y-4.5*\radius) circle (\radius);
    \foreach \i in {0,...,3}{
        \foreach \j in {0,...,3}{
            \draw[]
            (\x+8*\radius, \y-4.5*\radius+3*\i*\radius) --
            (\x+10*\radius, \y-4.5*\radius+3*\j*\radius);
        }
    }

    \filldraw[color=black, fill=gray!25, thick](\x+15*\radius,\y) circle (\radius);
    \draw[->,thick] (\x+16*\radius,\y) -- (\x+18*\radius,\y);
    \node[anchor=west] at
    (\x+18*\radius,\y) { \footnotesize Speed };
    \foreach \i in {0,...,3}{
        \draw[]
        (\x+12*\radius, \y-4.5*\radius+3*\i*\radius) --
        (\x+14*\radius, \y);
    }
}

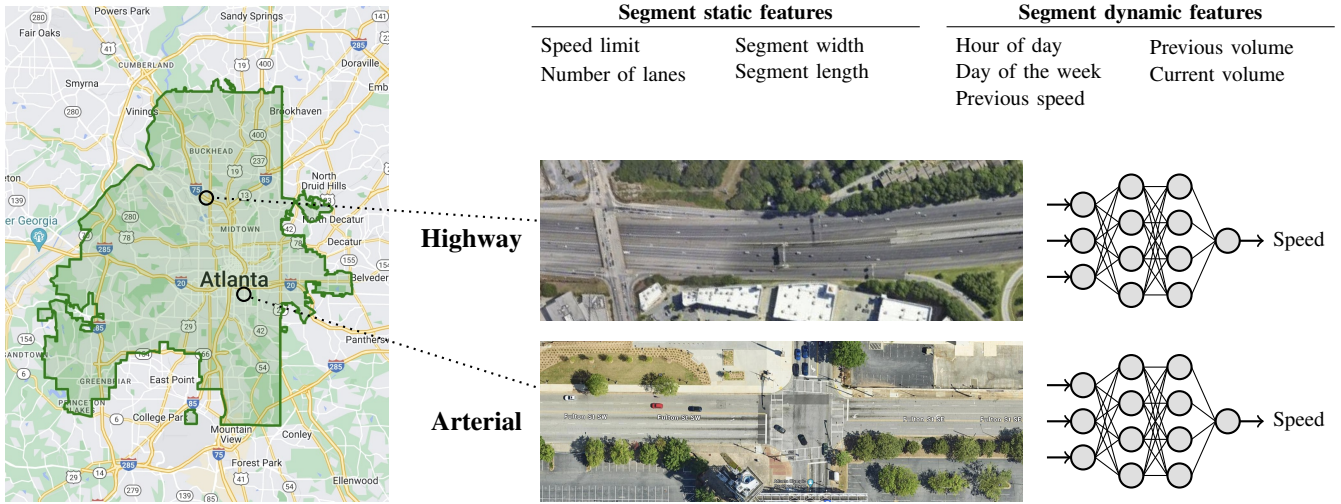
\begin{figure*}[th]
\centering
\begin{tikzpicture}
    \node[anchor=west] at
    (0,0) { \includegraphics[height=0.37\textwidth]{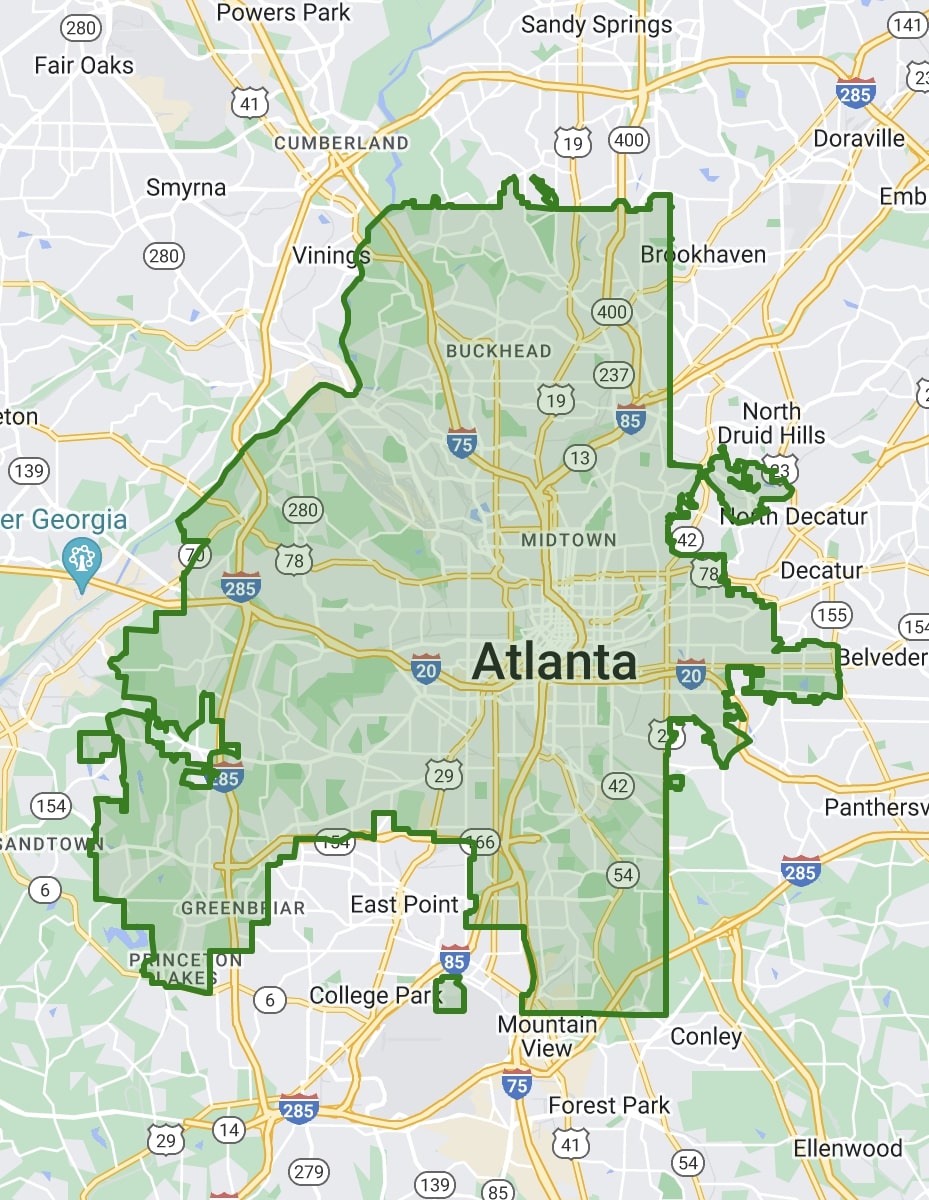}};

    \node[anchor=east] at
    (0.4\textwidth,0.01\textwidth) { \textbf{Highway} };
    \node[anchor=west] at
    (0.4\textwidth,0.01\textwidth) { \includegraphics[height=0.12\textwidth, width=0.36\textwidth]{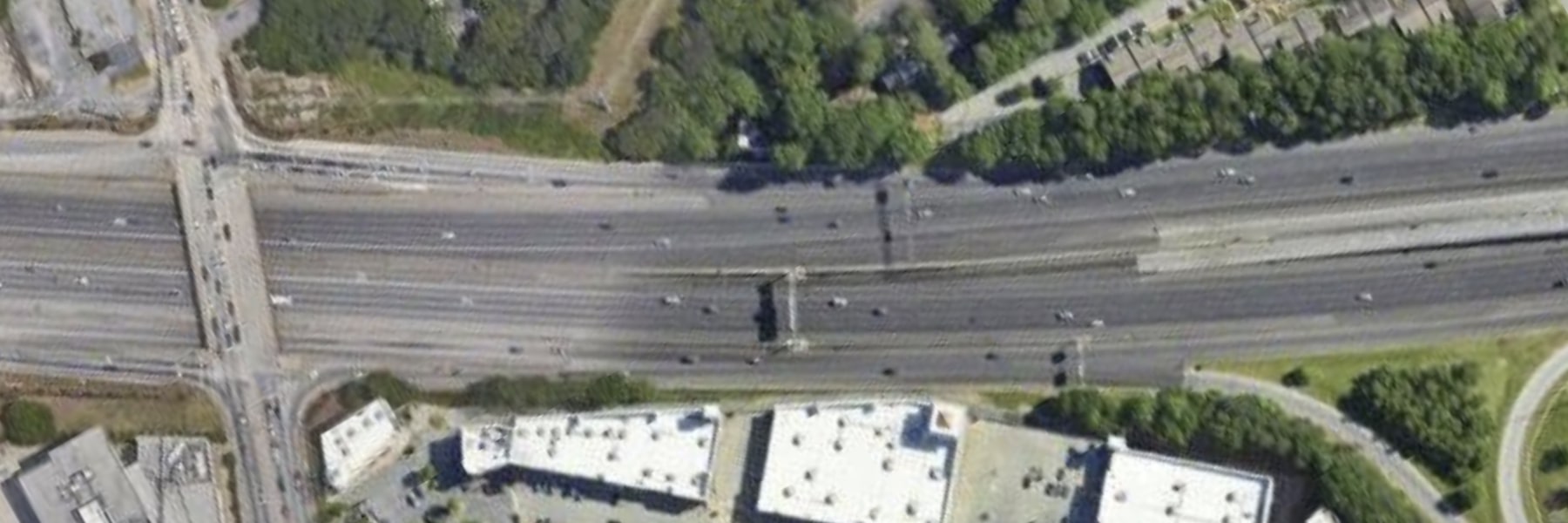}};
    \filldraw[color=black, fill=none, thick] (0.157\textwidth,0.042\textwidth) circle (0.005\textwidth);
    \draw[thick, dotted] (0.165\textwidth,0.042\textwidth) -- (0.405\textwidth,0.025\textwidth);
    \filldraw[color=black, fill=none, thick] (0.185\textwidth,-0.03\textwidth) circle (0.005\textwidth);
    \draw[thick, dotted] (0.1905\textwidth,-0.0325\textwidth) -- (0.405\textwidth,-0.1\textwidth);

    \node[anchor=east] at
    (0.4\textwidth,-0.125\textwidth) { \textbf{Arterial} };
    \node[anchor=west] at
    (0.4\textwidth,-0.125\textwidth) { \includegraphics[height=0.12\textwidth, width=0.36\textwidth]{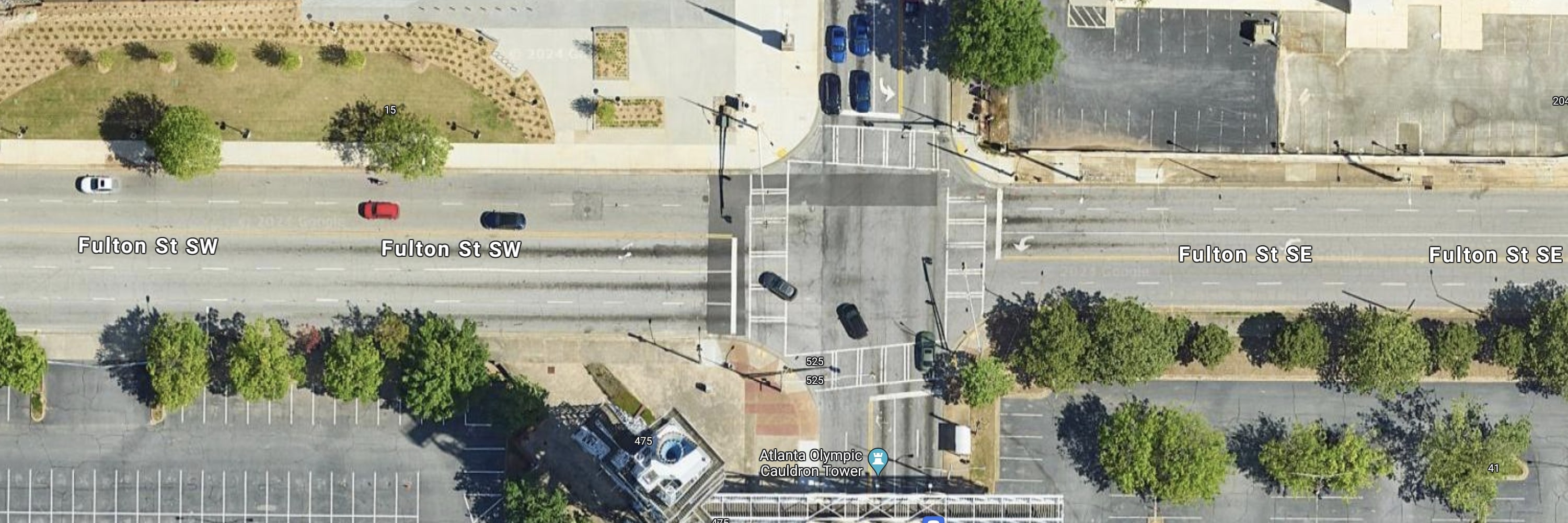}};

    \node[] at
    (0.545\textwidth,0.18\textwidth) { \footnotesize \textbf{Segment static features} };
    \draw[] (0.4\textwidth,0.17\textwidth) -- (0.69\textwidth,0.17\textwidth);

    \node[anchor=west] at (0.4\textwidth,0.155\textwidth) { \footnotesize Speed limit };
    \node[anchor=west] at (0.4\textwidth,0.135\textwidth) { \footnotesize Number of lanes };

    \node[anchor=west] at (0.545\textwidth,0.155\textwidth) { \footnotesize Segment width };
    \node[anchor=west] at (0.545\textwidth,0.135\textwidth) { \footnotesize Segment length };

    \node[] at
    (0.855\textwidth,0.18\textwidth) { \footnotesize \textbf{Segment dynamic features} };
    \draw[] (0.71\textwidth,0.17\textwidth) -- (1.0\textwidth,0.17\textwidth);

    \node[anchor=west] at (0.71\textwidth,0.155\textwidth) { \footnotesize Hour of day };
    \node[anchor=west] at (0.71\textwidth,0.135\textwidth) { \footnotesize Day of the week };
    \node[anchor=west] at (0.71\textwidth,0.115\textwidth) { \footnotesize Previous speed };

    \node[anchor=west] at (0.855\textwidth,0.155\textwidth) { \footnotesize Previous volume };
    \node[anchor=west] at (0.855\textwidth,0.135\textwidth) { \footnotesize Current volume };

    \neuralnetwork{0.785\textwidth}{0.01\textwidth}{0.009\textwidth}
    \neuralnetwork{0.785\textwidth}{-0.125\textwidth}{0.009\textwidth}
\end{tikzpicture}
\caption{An overview of our approach. We collect datapoints that capture static and dynamic properties of each segment within a given city. We then segregate datapoints by road type, in particular whether they are from a highway or arterial road. Finally, we train a separate neural network on each class of samples to estimate mean speed given current observed volume and other features. The overhead images are from Google Maps.}
\label{fig:overview}
\end{figure*}

Our key idea is to combine static and dynamic segment-level features, pool observed traffic data across segments, and learn congestion functions that generalize over them. Our approach could work on any form of the congestion function (e.g., estimating flow given density or density given speed); here we estimate space mean speed from observed partial flow. The static features capture segment geometry and configuration, e.g., length, width, speed limit etc., while the dynamic features include previous history of traffic data and temporal encodings to capture periodicity.

Figure~\ref{fig:overview} depicts the process we propose in this paper. We construct a single dataset of observed traffic data from all segments of a certain road type (highway or arterial) in an urban area. Each datapoint represents a single time interval on a single segment. The static and dynamic features are joined with the partial flow of that segment and time interval, while the target label is the observed space-mean speed. We then train a fully-connected feed-forward neural network on this regression dataset. Finally, we use the trained model as a congestion function that estimates segment-level speed from flow on held-out data for any segment in the set. Crucially, nothing in our approach would change if we observed total rather than partial flow. Thus, our approach could also have been used for identifying fundamental diagrams of traffic from appropriate data.

We conduct a detailed empirical evaluation of how our approach identifies flow-to-speed congestion functions and how it generalizes over segments. The cities in our dataset span different continents, traffic rules, and driving patterns.
The results demonstrate both strengths and limitations of our approach. For identifying mean speeds from flows on data from held-out time periods, our congestion functions compare favorably to a per-segment baseline on highway roads but not on arterial roads. On generalization, our approach is successful across cities and road types. When evaluated both on unobserved segments from the same city (with cross-validation) and on segments of a different city (with zero-shot transfer), the mean flow-to-speed estimation errors of our approach are comparable to when trained and tested on the same segments.

We contribute the following in this work:

\begin{itemize}
    \item A data-driven framework for identifying segment-level congestion functions that generalizes over segments and scales globally.
    \item A detailed empirical comparison with a segment-specific approach on multiple cities and road types.
    \item Strong results on generalization in both the same city and across cities through transfer learning.
    \item Initial results suggesting pooled models can predict traffic flow properties for individual segments.
\end{itemize}

\section{Related Work}
\label{sec:related}






\subsection{Identifying Fundamental Diagrams}
\label{sec:related-estimate-fdt}

The \emph{fundamental diagram} (FD) of traffic~\cite{haight1965mathematical} relates macroscopic traffic variables to each other. A rich history of research exists on proposing and using fundamental diagrams for traffic analysis and management~\cite{kuhne2008foundations}. 

Identifying fundamental diagrams typically involves gathering traffic data (either simulated or real-world) from a set of segments and then fitting a chosen functional form and noise parameters on a per-segment basis~\cite{courbon2011cross}. A standard practice is to use loop detectors, which measure flow and vehicle occupancy on a specific road segment. Then we estimate flow as a function of occupancy, which is a suitable proxy for true vehicle density. The recent benchmarking paper from~\citet{bramich2022fitting} comprehensively discusses this overall approach and its practical aspects. On a large multi-city loop detector dataset, it then compares a wide range of functional forms from the literature, from the early linear Greenshields model~\cite{greenshields1935study} to the penalized B-splines model of~\citet{sun2014data}.

Floating data from connected vehicles is an emerging source for fitting fundamental diagrams~\cite{ambuhl2017empirical}. The relevant work in this space are those that use machine learning models for the mapping between variables, e.g., gradient boosted decision trees~\cite{liu2019tailored} and long-short term memory neural networks~\cite{nam2020deep}. However, they implement and evaluate their approach on a small-scale per-segment basis only. Segment-specific approaches are useful and appropriate for particular high value roadways but can be hard to scale to all segments over large urban areas worldwide. 


\subsection{Traffic State Estimation}
\label{sec:related-tse}

For completeness, we briefly mention research on the related but distinct problem of traffic state estimation~\cite{wang2005real}. Formally, the traffic state estimation problem is defined as ``simultaneous estimation of flow, density, and speed on road segments with high spatio-temporal resolution, based on partially observed traffic data and a priori knowledge of traffic''~\cite{seo2017traffic}. In contrast to our problem of estimating one variable given observations of another, it estimates the values of the same variables at points in space and time without observations. Many approaches to traffic state estimation assume a given FD as a sub-component (though some data-driven ones do not~\cite{seo2015estimation}).

The literature on traffic state estimation, similar to that of FDs, works with different sources of input data and application scale~\cite{xing2022traffic} and approaches ranging from model-based~\cite{coifman2002estimating} to data-driven~\cite{duan2016efficient}. Much of the current state-of-the-art uses so-called \emph{physics-informed deep learning}, which~\citet{di2023physics} surveys comprehensively; they inject physics-based domain knowledge in a machine learning framework, typically through a surrogate loss function. A key finding of this work is that physics-informed learning works well in moderate data regimes, while data-driven learning is comparably good in big data regimes~\cite{huang2020physics}.

In this work, we took a purely data-driven approach without any physics-based assumptions. We are interested in exploring the strengths and limits of such data-driven learning by pooling information across segments. Designing a physics-informed deep learning approach in our context could be a promising area of future work.

\section{Problem Formulation}
\label{sec:problem}
We define a road segment as a physical stretch of road that represents travel between two intersections in one direction.
Our overall inputs are a set of road segments $\mathcal{S}$ and macroscopic traffic data from those segments over a time period T. Our traffic data consists of segment partial flow $q$ and (space-mean) speed $v$ per time interval, e.g., hourly. The speed estimates are for the entire roadway and not individual lanes. We denote this dataset as $(q_s^t, v_s^t)_{s \in \mathcal{S}, t = 1:T}$. 

The goal of identifying a congestion function (CF) is as follows: given $q_{s}^{t}$, the partial flow on segment $s$ at time interval $t$, estimate the corresponding space-mean speed, i.e., $\hat{v}_{s}^{t} = \theta\left(q_{s}^{t}\right)$, where $\theta$ is a parametric or nonparametric function representing the CF. Here, $\hat{v}$ indicates we are estimating the mean speed and we will evaluate a CF on its estimation accuracy on held-out data, i.e., $\mathrm{loss}(\hat{v}_{s}^{t}, v_{s}^{t}) \ \forall s \ \forall t$ for some appropriate loss metric. A high estimation accuracy suggests the CF will be useful for counterfactual analyses and simulations, which need a realistic estimate of speed given changing flow.

Typical model-based segment-level approaches assume a functional form and fit segment-specific parameters, i.e., $\hat{v}_{s}^{t} = \theta_{s}\left(q_{s}^{t}\right)$. For example, the classic Greenshields model assumes speed and density are linearly related, which leads to 
\begin{equation}
    q = v_{\text{ff}} \cdot \rho \cdot \left(1 - \frac{\rho}{\rho_{\text{crit}}}\right)
    \label{eq:greenshields}
\end{equation}
Here, the free parameters are free-flow speed $v_{\text{ff}}$ and critical density $\rho_{\text{crit}}$, both of which are estimated separately for each segment $s$, i.e., $\theta_s \equiv \{v_{\text{ff}, s}, \rho_{\text{crit}, s}\}$.

Our overall approach is to learn a \emph{single} data-driven congestion function over all given segments, i.e., $\hat{v}_{s}^{t} = \theta\left(q_{s}^{t}\right)$, where $\theta$ represents a supervised learning regression model. To train this data-driven congestion function, we will use additional segment-level features, defined in the next section.

\section{Learning Data-Driven Congestion Functions}
\label{sec:approach}

\subsection{Premise and Overview}
\label{sec:approach-overview}

Empirical flow-speed relationships are noisy and difficult to identify with analytical models~\cite{li2022fundamental,bramich2022fitting}. As with other regression tasks, using features (both static segment attributes and time-varying dynamic signals) that plausibly influence the output variable can help improve estimation error. 
We could have just used features to learn a separate CF for every individual segment, but this approach is also unsatisfying. In principle, if the features influence traffic dynamics, then segments and time-intervals with similar features should have similar flow-to-speed mappings, and we should be able to learn a single CF across segments. Such a CF can both work well for segments with many observations and generalize to roadways with little or no data (which a per-segment CF, whether data-driven or analytical, cannot). In practice, maintaining and updating separate models for every individual roadway in a large urban area could be challenging.

For each segment $s \in \mathcal{S}$ and time interval $t = 1:T$, we create a static feature vector $\phi_s$ and a dynamic feature vector $\Phi_{s}^{t}$. We construct our regression dataset by pooling across all segments and time intervals. Each input example is the joined feature vectors and the observed partial flow, $\left(\phi_s, \Phi_{s}^{t}, q_{s}^{t}\right)$, and the corresponding label is the space-mean speed ${v}_{s}^{t}$. We train a machine learning model (whose parameters are encoded by $\theta$) on this dataset to produce a speed estimate $\hat{v}_{s}^{t}$, supervised by some loss function $\mathrm{loss}\left({v}_{s}^{t}, \hat{v}_{s}^{t}\right)$. At inference or evaluation time, given held-out partial flow $q_{s}^{t}$ from segment $s$ and time period $t$, we first construct the input vector $\left(\phi_s, \Phi_{s}^{t}, q_{s}^{t}\right)$ and then estimate $\hat{v}_{s}^{t} = \theta\left(q_{s}^{t}\right) \equiv \theta\left(\phi_s, \Phi_{s}^{t}, q_{s}^{t}\right)$.

\subsection{Feature Description}
\noindent
\textbf{Static Features}: These are attributes of the segment that could influence the traffic variables and thus help our approach generalize across segments. They are not strictly static in that their values may change in the long-term. But for any specific dataset, these features do not vary. We use the following:

\begin{itemize}
    \item \emph{Length}: The length of the segment's polyline geometry, in meters.
    \item \emph{Lanes}: The number of individual lanes along the segment direction.
    \item \emph{Width}: The sum of widths of the segment lanes, in meters.
    \item \emph{Speed Limit}: The posted speed limit of the segment, in meters per second.
\end{itemize}

\noindent
\textbf{Dynamic Features}: These are time-varying quantities that we expect will help to model the flow-speed relationship better, both for a given segment and when pooling across segments. We use the following:

\begin{itemize}
    \item \emph{Temporal Encoding}: Sine-cosine encoding of hour-of-day and day-of-week of the measurements is a commonly used method to capture periodicity in the signals.
    \item \emph{Normalized Flow}: Flow divided by the product of segment length and number of lanes, i.e., $\nicefrac{q_s^t}{\text{len}(s)*\text{lane}(s)}$
    \item \emph{Previous Measurements}: We use the flow and speed from the previous time-interval when available, i.e., $q_s^{t-1}, v_s^{t-1}$. Traffic variables in one time interval can depend strongly on the previous one.
\end{itemize}

\subsection{Cluster by Priority}

The segment attribute that most influences the flow-speed relationship is the road priority (also called class or type); the importance of the road for carrying traffic, whether it allows through traffic, and how it is accessed. The exact categories can vary by country and region. For convenience, we consider two that are easy to distinguish:

\begin{enumerate}
    \item \emph{Highway}, i.e., major roads that connect regions. They may also be partially or fully access-controlled.
    \item \emph{Arterial}, i.e., moderate-to-high capacity roads that either carry traffic between towns or neighborhoods, or from smaller local roads to larger ones.
\end{enumerate}

Each road priority has fundamentally different traffic characteristics and attribute distributions; between-group variation is larger than in-group variation. We assume that the segments of one priority will not help learn CFs for another. Thus, for each urban area, we will \emph{cluster segments by priority and learn a separate data-driven CF for each priority}, i.e., two CFs per urban area.

\subsection{Machine Learning Pipeline}

We now have all the ingredients for applying a typical supervised machine learning pipeline.

\subsubsection{Datasets}
Three overall parameters determine any particular dataset: the set of segments (of given area and priority), the time resolution, and the date range. We join static features for all segments with the traffic data and dynamic features of each segment per time interval over the date range, e.g., every hour over the past month. We filter data based on static features, e.g, minimum segment length, and dynamic, e.g., by speed or time of day. Since we care about predictive utility on held-out data, the validation and testing datasets are constructed from non-overlapping date ranges, e.g., train on data from March and validate and test on the first and second weeks of April respectively.~\Cref{sec:experiments} will mention all relevant specifics for our reported results.

\subsubsection{Model}

We use a fully-connected feed-forward neural network as our model~\cite{lecun2015deep}. The first layer normalizes each input feature to zero mean and unit standard deviation across the dataset. For activation functions, we use the exponential linear unit  for all but the final layer~\cite{clevert2015fast}. The output layer uses the sigmoid activation $\sigma(x) = \frac{1}{1 + e^{-x}} \in (0, 1)$; as we explain shortly, our prediction target is the inverse of the mean speed, which is always greater than 1 in our dataset.

\subsubsection{Loss Function}

In general, congestion functions are more important when road segments are becoming congested, i.e., when mean speeds are low. A simple trick to encourage better estimation in those regimes is to try and predict inverse speed, i.e., $\frac{1}{\hat{v}_{s}^{t}}$. If we then use the standard $L_2$-squared loss (omitting segment $s$ and time $t$ for readability),

\begin{equation}
    \mathrm{loss}\left(\hat{v}, v\right) = \frac{1}{\hat{v}^2} - \frac{1}{v^2}, \ \ (v, \hat{v} > \SI{1}{\metre\per\second})
\end{equation}
we can penalize worse predictions in congested regimes. Since we filter implausible cases where $v < \SI{1}{\metre\per\second}$, we can assume $\frac{1}{\hat{v}} \in (0,1)$, which lets us use the sigmoid output activation.

\section{Experiments}
\label{sec:experiments}

\subsection{Large-Scale Dataset}
\label{sec:experiments-data}

Our dataset covers major global metropolitan areas, with different road network structure and traffic patterns. Evaluating our approach on such a large scale emphasizes how broadly applicable it is. We use static segment features and aggregated and anonymized space-mean speeds and vehicle counts from the Google Maps road network data and driving trends. ~\Cref{tab:city_details} has some relevant summary statistics. In principle, our approach can apply to any urban area in the world; here we show selected results due to space constraints.

\begin{table}[t]
    \caption{Overview of the global scale of our dataset. Each quantity is noted as a multiple of one thousand.}
    \centering
    \begin{tabular}{@{} lcrrr @{}}
        \toprule
        Metro Area && \#Highway Segs & \#Arterial Segs & Area km$^2$ \\
        \midrule
        Atlanta && $45\text{e}3$  & $89\text{e}3$ & $66\text{e}3$ \\
        Los Angeles && $63\text{e}3$ & $20\text{e}3$ & $47\text{e}3$ \\
        Munich  && $12\text{e}3$ & $43\text{e}3$ & $54\text{e}3$\\
        London  && $3\text{e}3$ & $98\text{e}3$ & $9\text{e}3$\\
        Dubai  && $3\text{e}3$ & $17\text{e}3$ & $6\text{e}3$\\
        Osaka  && $27\text{e}3$ & $32\text{e}3$ & $11\text{e}3$\\
        \bottomrule
    \end{tabular}
    \label{tab:city_details}
\end{table}

\begin{table}
    \centering
    \caption{Our neural network setup for all experiments.}
    \begin{tabular}{@{} lcr @{}}
    \toprule
    Configuration && Value \\
    \midrule
    Architecture && (Input, 16, 8, 4, 8, 16, 1) \\
    Activations && (elu, elu, $\ldots$, sigmoid) \\
    Batch Size && 256 \\
    Shuffle Buffer && 10000 \\
    Epochs && 30 \\
    \bottomrule
    \end{tabular}
    \label{tab:nn_config}
\end{table}

Our flow and speed measurements are collected every hour. We experimented with finer resolutions, i.e., every 20 and 30 minutes, and found similar overall performance but noisier observations. Hence, we stick with hourly here. We used all days of the week and 7am to 10pm each day. We filtered segments shorter than $20$ meters and any hourly examples where speed was outside \SIrange{1}{45}{\meter\per\second}.

Each dataset is created for the set of all segments of a metro area and priority (e.g., Atlanta-Highway, Munich-Arterial and so on) and time range. For every segment set, the training dataset covers about 5 weeks, the validation dataset the following week, and the testing dataset the week after that. Here we are interested in capturing short-term flow-speed relationships rather than long-term traffic patterns.

\begin{table*}[th]
    \caption{Aggregate mean absolute error (MAE) of our approach with one model for all segments (All-Seg ML), compared against the BPR baseline with one model per segment (Per-Seg BPR). To emphasize the advantage of our approach, we also report its MAE on the segments for which the BPR approach does not work at all (BPR-No-Fit). Overall, our approach has generally lower MAE on highway segments but much more room to improve on arterials.}
    \centering
    \begin{tabular}{@{} lcrrrcccrrr @{}}
      \toprule
      & \phantom{a} & \multicolumn{3}{c}{Highway MAE (\SI{}{\meter/\second})} & \phantom{a} & \phantom{a} & \phantom{a} & \multicolumn{3}{c}{Arterial MAE (\SI{}{\meter/\second})}\\
      \cmidrule{3-5} \cmidrule{9-11} City && All-Seg ML & Per-Seg BPR & ML on BPR-No-Fit &&&& All-Seg ML & Per-Seg BPR & ML on BPR-No-Fit \\
      \midrule
      Los Angeles   && 1.60 & 1.86 & 0.87 &&&& 1.27 & 0.73 & 2.69  \\
      Atlanta       && 1.11 & 1.09 & 1.03 &&&& 1.22 & 0.78 & 1.97  \\
      London        && 1.08 & 0.99 & 0.60 &&&& 1.03 & 0.57 & 1.63  \\
      Munich        && 0.80 & 0.79 & 0.50 &&&& 1.15 & 0.68 & 2.17  \\
      Dubai         && 1.05 & 1.40 & 0.55 &&&& 1.04 & 0.77 & 1.07 \\
      Osaka         && 1.12 & 0.85 & 1.02 &&&& 1.14 & 0.68 & 1.62 \\
      \bottomrule
    \end{tabular}%
    \label{tab:aggregate_metrics}
\end{table*}

\subsection{Implementation Details}
\label{sec:experiments-impl}

~\Cref{tab:nn_config} mentions the settings of the neural network trained on each dataset. We used a black-box hyperparameter optimization scheme for the chosen values. The best setting varied slightly for each dataset; but for convenience and simplicity we used the same configuration for all cases. All experiments were run using Python and Tensorflow~\cite{abadi2016tensorflow}.

\subsection{Evaluations and Results}
\label{sec:experiments-results}

We examined two properties of our data-driven congestion function:  downstream predictive utility and ability to generalize. For predictive utility, we evaluated it on a held-out week of test data over the same set of segments. We compared against a per-segment baseline to put results in context. For generalization, we trained on data from a portion of segments and evaluated it on data from the remaining segments. We also tested a trained model from one city on data from another city, i.e., zero-shot transfer learning.

In all cases, we report one of two standard performance metrics for regression: mean absolute error (\emph{MAE}), i.e., $\frac{1}{N} \sum \lVert v_i - \hat{v}_i \rVert$ and mean absolute percentage error (\emph{MAPE}), i.e., $\frac{1}{N} \sum \lVert \frac{v_i - \hat{v}_i}{v_i} \rVert$, where $N$ is sample size. Lower is better for both metrics. All error estimates are computed over test datasets where $N$ is in the tens of millions. Therefore, the standard error of the mean, which is  proportional to $\frac{1}{\sqrt{N}}$, is negligible (at least three orders of magnitude smaller than the mean) and we do not report it.

\subsubsection{Baseline: Per-Segment CF}

We compare the predictive utility of our approach against a commonly used parametric model. We fit samples from each segment $s$ to the \textit{Bureau of Public Roads} (BPR) function~\cite{us1964traffic}, which models the relationship between speeds $v_s$ and densities $\rho_s$ as
\begin{equation}
\frac{1}{v_s} =
\begin{cases}
\frac{1}{v_\text{FF}} & \text{if } \rho_s < \rho_\text{crit} \\
\frac{1}{v_\text{FF}} + c \cdot \left( \frac{\rho_s}{\rho_\text{crit}} - 1 \right) ^ p & \text{otherwise}
\end{cases}
\label{eq:bpr}
\end{equation}
where $\rho_\text{crit}$ is the critical density of the segment, $v_\text{FF}$ is its free-flow speed, and $c$ and $p$ and linear and polynomial scaling terms respectively. These parameters are fit to provided samples for each segment in the training set using a Trust Region Reflective algorithm implemented within SciPy~\cite{virtanen2020scipy}.

\subsubsection{Aggregate Error Metrics}

For each city and priority, we take the trained CF from roughly 5 weeks of data and evaluate aggregate error on the test week. We do so for 5 different random seeds and report the best run in each case. The BPR baseline has the same train-test protocol.

~\Cref{tab:aggregate_metrics} reports the aggregate MAE (in \SI{}{\meter\per\second}) of our approach (\emph{All-Seg ML}) and the baseline (\emph{Per-Seg BPR}) for all datasets. Each set has a non-trivial proportion of segments with too few datapoints for which to fit the BPR baseline. This proportion varies from \numrange[range-phrase = --]{5}{10}\% for highways and \numrange[range-phrase = --]{10}{20}\% for arterials. Of course, such segments are not an issue for our approach.  To further emphasize its scalability, we split the aggregate test-set MAE of our approach into the MAE on segments with BPR fits, and the MAE on segments without BPR fits (\emph{ML on BPR-No-Fit}).

On highway roads, our data-driven single CF performs strongly against the per-segment BPR; it has comparable MAE on the segments with BPR fit and similar or even lower MAE on the remaining segments (except in Osaka). However, on arterial roads the single CF struggles to match the baseline. Arterials could have a wider range of traffic dynamics, making it difficult to fit a single function to their data. Our approach might need more segment features and more training data to perform better compared to the baseline. The comparison with per-segment BPR is also less fair in general on arterials, as there is a higher proportion of segments without enough data for a BPR fit in the first place.

 A quick comment on the raw loss of predicting inverse speed itself: the mean squared error ranges from \numrange[range-phrase = --]{3}{7} units on highways and \numrange[range-phrase = --]{3}{4} units on arterials.

\subsubsection{Disaggregated Error Metrics}

Aggregate metrics capture overall performance, but underlying traffic conditions vary throughout the dataset and influence prediction error. A scalable segment-agnostic way to disaggregate the error is by \emph{quartiles of normalized speed}, i.e., the mean speed of the datapoint divided by the speed limit of the segment. The first quartile represents maximum congestion (a speed to limit ratio of less than $0.3$). The second and third quartiles typically cover transitions between congested and free-flow periods (a ratio of between $0.3$ and $0.6$), while the last quartile represents more free-flow traffic (higher than $0.6)$. 

~\Cref{fig:disagg_mape_metrics} shows the disaggregated MAE per normalized speed quartile on highway segments of selected cities: Atlanta, Munich, and Dubai. The baseline has lower error in the first quartile, while our approach is consistently better in the remaining quartiles. This suggests the traffic behavior in maximally, congested periods varies more widely across segments, making it harder to identify a single CF across all of them.

\begin{figure*}[t]
    \centering
    \begin{subfigure}{0.32\textwidth}
    \includegraphics[width=\columnwidth]{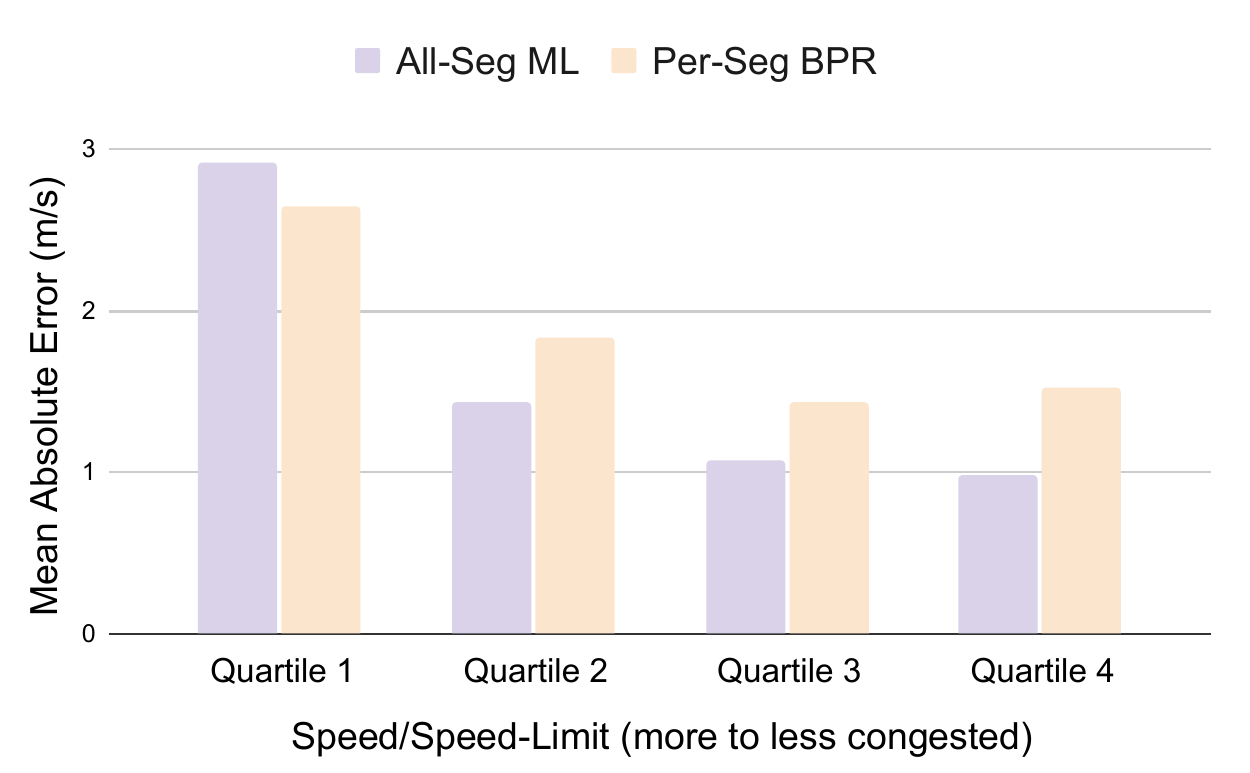}
    \caption{Los Angeles Highway}
    \end{subfigure}
    \hfill
    \begin{subfigure}{0.32\textwidth}
    \includegraphics[width=\columnwidth]{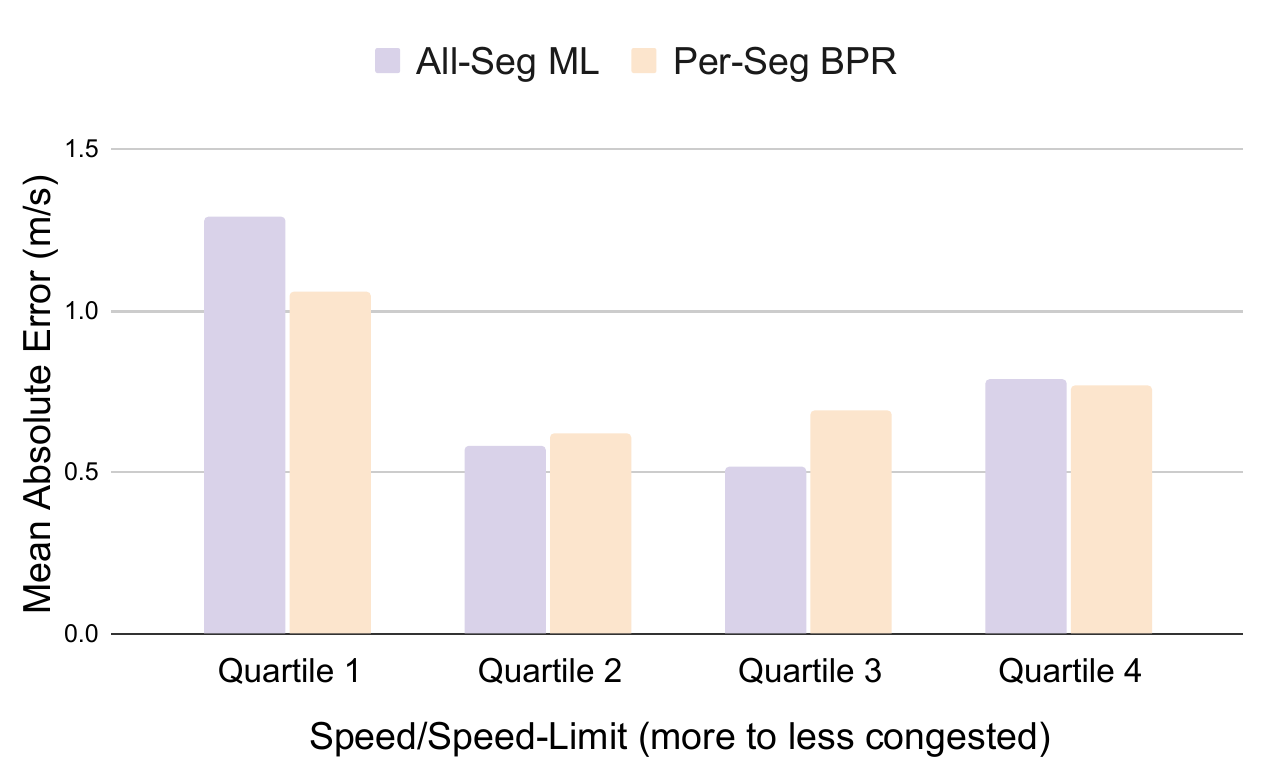}
    \caption{Munich Highway}
    \end{subfigure}
    \hfill
    \begin{subfigure}{0.32\textwidth}
    \includegraphics[width=\columnwidth]{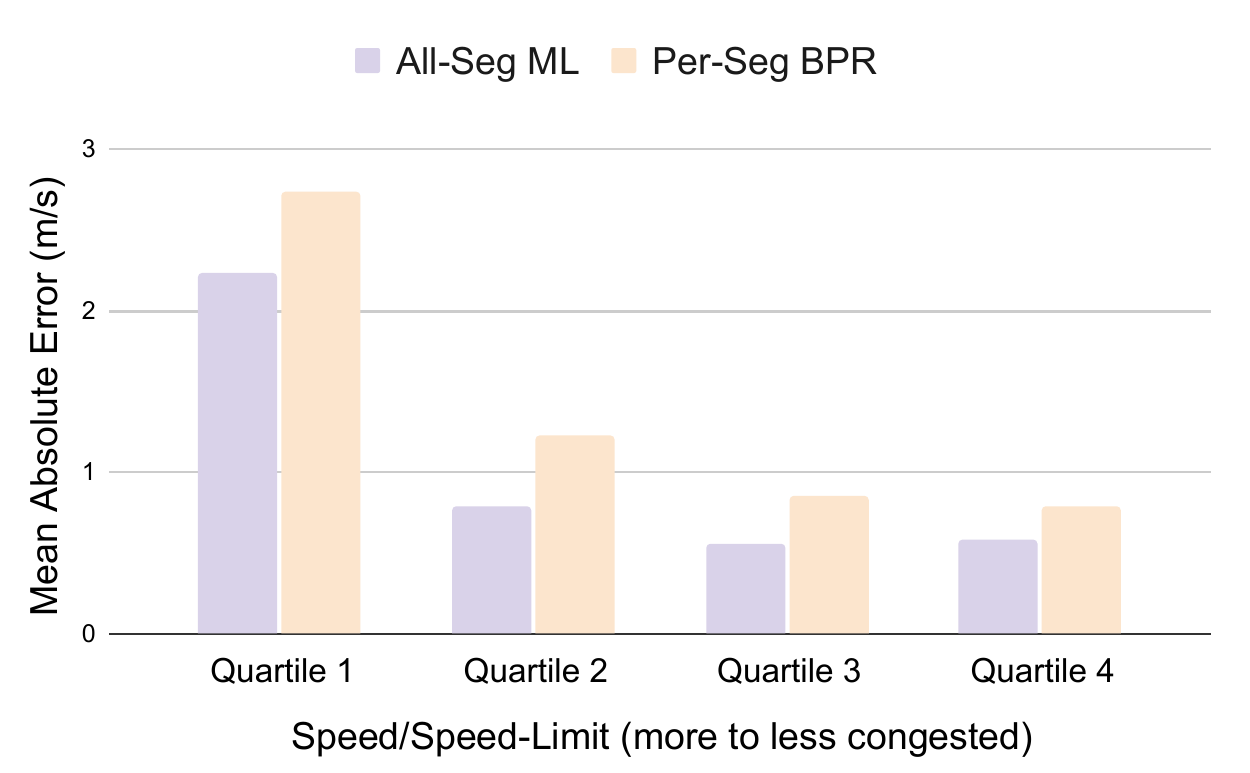}
    \caption{Dubai Highway}
    \end{subfigure}
    \caption{The MAEs from~\Cref{tab:aggregate_metrics}, disaggregated by quartiles of normalized speed, i.e., speed divided by segment speed limit. We focus on highway segments of three representative cities here. Our approach has lower MAE in periods of free-flow and transition between congestion and free-flow, while the baseline is better during periods of most congestion.}
    \label{fig:disagg_mape_metrics}
\end{figure*}

\subsubsection{Generalizing to unobserved segments}

Our approach has already shown it can generalize, since we learn a single CF across a set of segments using static and dynamic features. Now we test the trained CFs on explicitly unobserved segments. We divide a given segment set roughly 80-20 into train and test sets. We collect roughly 5 weeks of data for both sets of segments, then we train on the former and evaluate on the latter. To be more robust to the choice of 80-20 split, we do 5-fold cross validation, which divides the segment set into 5 \textit{folds} of roughly $20\%$ each. In the first run, the first fold is for testing and the last four are for training. In the second run, the second fold is for testing and the other four for training. We repeat this accordingly for 5 runs and report the median metric.

~\Cref{fig:cross_val_mape} compares cross-validation mean absolute percentage error (MAPE) with the MAPE from training and testing the CF on the same set of segments (for selected cities and both priorities). In all cases, the numbers are comparable, i.e., the points are close to the $x=y$ line. This findings suggests that our single CF can work as well on explicitly unobserved segments as it does on the training segments.

\begin{figure}
    \centering
    \includegraphics[width=0.8\columnwidth]{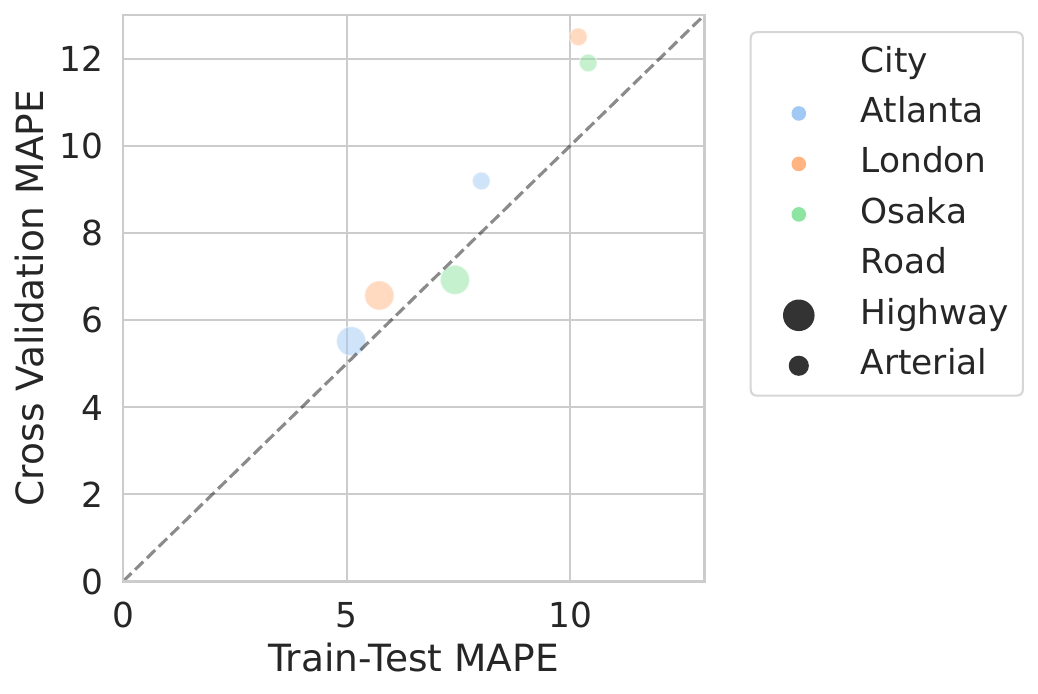}
    \caption{Our approach generalizes well to unseen segments. We train it on 80\% of the segments for each dataset, and evaluate it on the remaining 20\% for the same time period. We do this through 5-fold cross-validation and report the median MAPE of the 5 folds here (y-axis) against the MAPE of training and testing on the same set of segments (x-axis).}
    \label{fig:cross_val_mape}
\end{figure}

\subsubsection{Zero-shot Transfer between Cities}

Now we study another generalization question: can a feature-based data-driven CF transfer across entire cities? Specifically, we look at zero-shot learning, a special case of transfer learning~\cite{pan2009survey}. Here, that involves taking a CF trained on segments of a city and priority, and evaluating it directly on the test data from another city (and the same priority). ~\Cref{fig:zero_shot_metrics} shows the MAE of various city-to-city transfer learning evaluations, separated by priority. As with cross-validation, the transfer metrics are compared against the metrics from training and testing on the downstream city.  The results are intuitive yet insightful. Cities in the USA and Europe (here, Atlanta, Los Angeles, Munich, London) tend to transfer well between each other; the points are closer to the $x=y$ line. On the other hand, a city from Asia (here, Osaka) transfers less well to both Atlanta and London, suggesting the limits of transfer across the globe.

\begin{figure}
    \centering
    \includegraphics[width=0.85\columnwidth]{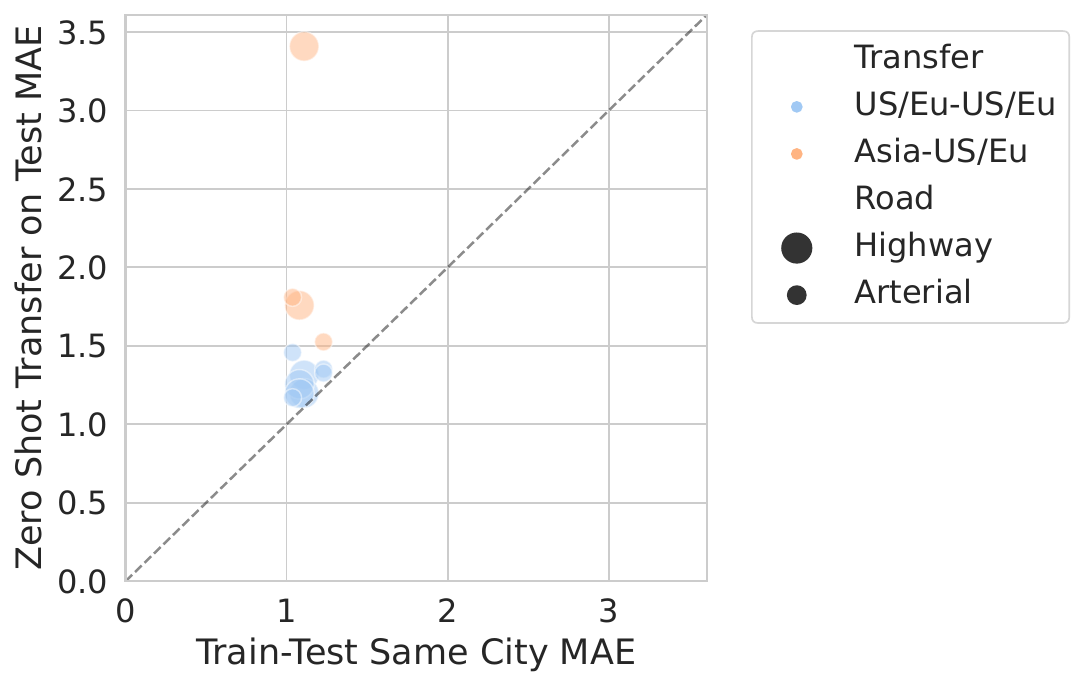}
    \caption{For selected pairs of cities, we plot the MAE of training on the first city and testing on the second (y-axis), against the MAE of training and testing on the second (x-axis). For pairs where both cities are from USA or Europe, the points are closer to the x=y line, suggesting better transfer possibly due to similar traffic patterns. The opposite is true when one of cities is from Asia.}
    \label{fig:zero_shot_metrics}
\end{figure}

\subsection{Deriving segment-specific traffic flow properties}

To further explore the downstream utility of our framework, we now examine if our generated pooled models can predict relevant traffic flow properties of a segment. In particular, we study the models' ability to predict the critical density of a segment, or the density at which a segment transitions from free-flow to congested conditions. Numerically, we define the predicted critical density of a segment as the density at which predicted flows are maximal.

We compute the critical density from our data-driven congestion function through its predicted speed estimates on samples in the held-out evaluation set. We then compare the predicted critical densities of our models with ground truth measurements extracted from the complete dataset for a given segment (both train and test sets) as well as predicted critical densities from the per-segment BPR models. For ground truth, we use a similar definition of critical density as we used for the pooled model, specifically computing the critical densities at the density at which ground truth partial flows are maximal. For per-segment BPR models, we exploit their parametric nature and set the predicted critical densities to the $\rho_\text{crit}$ term from ~\Cref{eq:bpr}.

Figure~\ref{fig:inferring_properties} depicts the distribution of predicted critical densities for segments within the city of Atlanta. Cross-relational differences in estimated critical densities by each pair of approaches can be found in Table~\ref{tab:properties-inference-comparison}. Seen here, the three approaches provide estimates that are relatively similar in magnitude, with estimates by the per-segment model often falling on the lower end of those provided by the other two approaches. These finding suggest that the pooled ML model can predict useful properties of a segment while only using segment attributes to distinguish between them.

\begin{figure}
    \centering
    \includegraphics[height=0.3\textwidth]{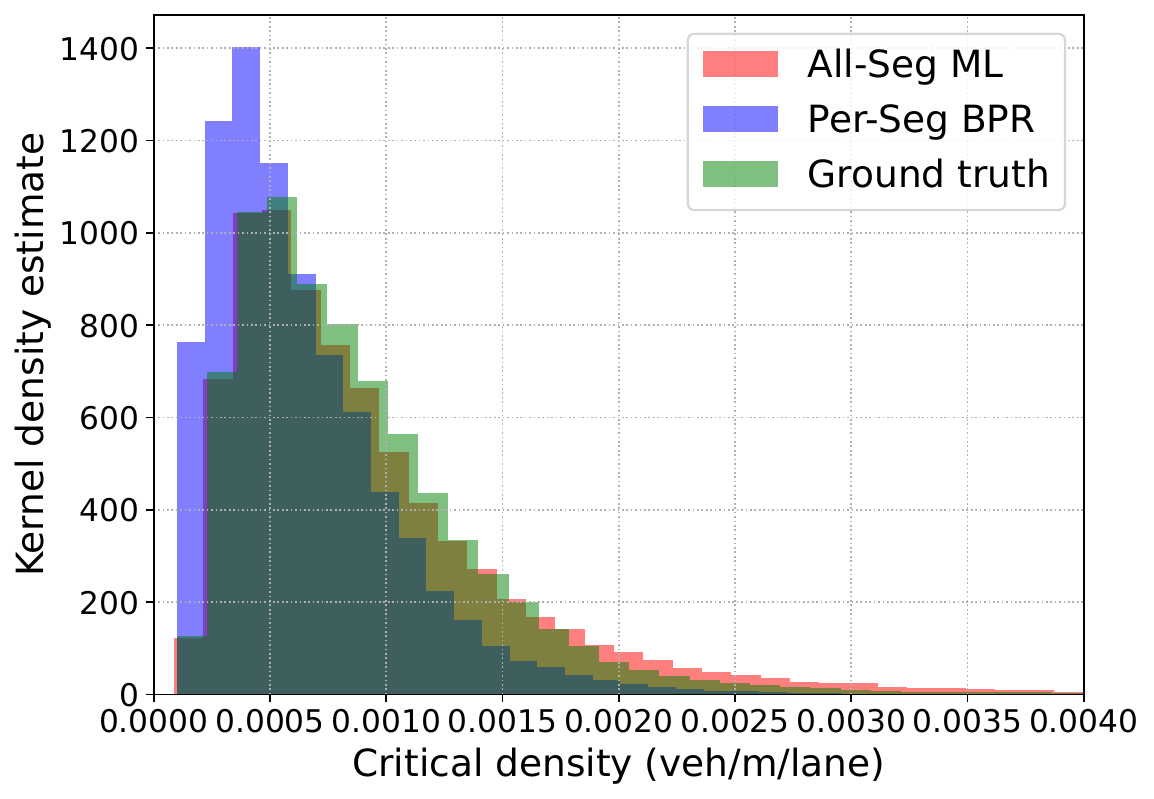}
    \caption{Distributions of predicted critical densities from different methods.}
    \label{fig:inferring_properties}
\end{figure}

\begin{table}[t]
    \centering
    \begin{tabular}{lllll}
    \toprule
    \textbf{Road category} & \textbf{Comparison} & \textbf{MAE} & \textbf{MAPE} \\ \midrule
    \multirow{3}{*}{Highway} & Ground truth / Per-Seg BPR & 3.4e-4 & 0.394 \\
    & Ground truth / All-Seg ML  & 2.2e-4 & 0.229 \\
    & All-Seg ML / Per-Seg BPR  & 5e-4 & 0.465 \\ \midrule
    \multirow{3}{*}{Arterial} & Ground truth / Per-Seg BPR  & 5.8e-4 & 0.599 \\
    & Ground truth / All-Seg ML  & 1.4e-4 & 0.137 \\
    & All-Seg ML / Per-Seg BPR  & 6.2e-4 & 0.624 \\ \bottomrule
    \end{tabular}
    \caption{Cross-relational errors between each of the ground truth, all-seg BPR, and per-seg ML estimates of critical densities for segments with some level of congestion.}
    \label{tab:properties-inference-comparison}
\end{table}

\section{Discussion}
\label{sec:discussion}

A solution that can scale and generalize globally like ours typically has context-dependent trade-offs against one that is tuned for each individual roadway, like the BPR baseline or any classical per-segment approach. If our segment-pooling data-driven approach can identify CFs roughly as well as the segment-specific model-driven baseline on the same roadways, while also generalizing to unseen roadways in a way that the baseline cannot, that would be a win-win. As ~\Cref{tab:aggregate_metrics} and ~\Cref{fig:disagg_mape_metrics} show, \emph{we currently get a win-win on highway segments}. On arterial segments, we have much more room to improve. But the model-driven segment-specific approaches have been worked on for decades~\cite{kuhne2008foundations}, while the approach we present here is a first attempt. We do not focus on getting the best possible machine learning result here; our features are not exhaustive and our architecture and hyperparameters are simple. We spent the bulk of our effort on evaluating the strengths and limitations of our idea as compared to the baseline, at a larger scale than any prior work. Future work could iterate indefinitely on better architectures and pipelines, e.g, using physics-informed constraints, graph neural networks, and more sophisticated segment clustering and pooling. 

The generalization experiments support our premise that \emph{we can identify CFs of roadways without fitting a specific set of parameters for each one}. Here too, state-of-the-art techniques from scalable machine learning, including finetuning on samples from the unseen segments or cities, and meta-learning of CFs, could further improve the results we already have. On the other hand, for applications where only a few high-value roadways matter, a per-segment model-driven approach may continue to be the most suitable. We view our work as adding to the set of techniques available to the transportation research community and enabling them to make a wider impact than they currently do.

\section{Conclusion}
\label{sec:conclusion}

This paper described and evaluated a scalable data-driven approach for identifying segment-level traffic congestion functions. Our key idea was to pool data across all roadways of the same type, and use static and dynamic features to learn CFs that generalize across them. Results for both identification performance and generalization were promising. Our approach had better speed estimation error than the segment-specific baseline on highways, both overall and in free-flow and transition periods when disaggregated by speed quartile. It also generalized successfully to unobserved segments from the same city and to other cities with similar traffic patterns. The global scale of our experiments, covering most road segments of multiple major cities worldwide, emphasize how widely applicable our work can be.

Future research could improve several aspects of our framework. Our segment features and highway/arterial distinctions were simple. More complex features (e.g., stop signs, traffic lights, neighboring segments) and better segment clustering techniques could improve overall estimation accuracy, especially on tricky arterial roads. The machine learning pipeline could include more inductive biases, such as the graphical structure of the road network, e.g., through graph neural networks, or physics-informed constraints. 

We could extend the approach to solve related but distinct formulations of the overall task, e.g., inferring other traffic properties on the roadways or forecasting congestion in the near future. Evaluating our approach on a large open-source dataset, such as the one from~\citet{neun2023metropolitan} would further prove its value. Finally, scalable learning-based congestion functions could enable a new approach to citywide mesoscopic simulations that use such models in the loop.

\section*{Acknowledgments}
The authors would like to thank Pranjal Awasthi, Chun-Ta Lu, and Andrew Tomkins (all from Google Research) for their advice and comments.

\bibliographystyle{IEEEtranN}
\bibliography{references}

\begin{thebibliography}{29}
\providecommand{\natexlab}[1]{#1}
\providecommand{\url}[1]{#1}
\csname url@samestyle\endcsname
\providecommand{\newblock}{\relax}
\providecommand{\bibinfo}[2]{#2}
\providecommand{\BIBentrySTDinterwordspacing}{\spaceskip=0pt\relax}
\providecommand{\BIBentryALTinterwordstretchfactor}{4}
\providecommand{\BIBentryALTinterwordspacing}{\spaceskip=\fontdimen2\font plus
\BIBentryALTinterwordstretchfactor\fontdimen3\font minus
  \fontdimen4\font\relax}
\providecommand{\BIBforeignlanguage}[2]{{%
\expandafter\ifx\csname l@#1\endcsname\relax
\typeout{** WARNING: IEEEtranN.bst: No hyphenation pattern has been}%
\typeout{** loaded for the language `#1'. Using the pattern for}%
\typeout{** the default language instead.}%
\else
\language=\csname l@#1\endcsname
\fi
#2}}
\providecommand{\BIBdecl}{\relax}
\BIBdecl

\bibitem[Ljung(1998)]{ljung1998system}
L.~Ljung, ``System identification,'' in \emph{Signal analysis and
  prediction}.\hskip 1em plus 0.5em minus 0.4em\relax Springer, 1998, pp.
  163--173.

\bibitem[Deng et~al.(2021)Deng, Zhang, and Shen]{deng2021systematic}
T.~Deng, K.~Zhang, and Z.-J.~M. Shen, ``A systematic review of a digital twin
  city: A new pattern of urban governance toward smart cities,'' \emph{Journal
  of Management Science and Engineering}, vol.~6, no.~2, pp. 125--134, 2021.

\bibitem[Daganzo(1997)]{daganzo1997fundamentals}
C.~F. Daganzo, \emph{Fundamentals of transportation and traffic
  operations}.\hskip 1em plus 0.5em minus 0.4em\relax Emerald Group Publishing
  Limited, 1997.

\bibitem[Bramich et~al.(2022)Bramich, Men{\'e}ndez, and
  Amb{\"u}hl]{bramich2022fitting}
D.~M. Bramich, M.~Men{\'e}ndez, and L.~Amb{\"u}hl, ``Fitting empirical
  fundamental diagrams of road traffic: A comprehensive review and comparison
  of models using an extensive data set,'' \emph{IEEE Transactions on
  Intelligent Transportation Systems}, vol.~23, no.~9, pp. 14\,104--14\,127,
  2022.

\bibitem[Treiber et~al.(2013)Treiber, Kesting, Treiber, and
  Kesting]{treiber2013trajectory}
M.~Treiber, A.~Kesting, M.~Treiber, and A.~Kesting, ``Trajectory and
  floating-car data,'' \emph{Traffic Flow Dynamics: Data, Models and
  Simulation}, pp. 7--12, 2013.

\bibitem[Haight(1965)]{haight1965mathematical}
F.~A. Haight, \emph{Mathematical theories of traffic flow}, 1965.

\bibitem[Kuhne(2008)]{kuhne2008foundations}
R.~D. Kuhne, ``Foundations of traffic flow theory i: Greenshields'
  legacy--highway traffic,'' in \emph{Symposium on the Fundamental Diagram: 75
  Years (Greenshields 75 Symposium) Transportation Research Board}, 2008.

\bibitem[Courbon and Leclercq(2011)]{courbon2011cross}
T.~Courbon and L.~Leclercq, ``Cross-comparison of macroscopic fundamental
  diagram estimation methods,'' \emph{Procedia-Social and Behavioral Sciences},
  vol.~20, pp. 417--426, 2011.

\bibitem[Greenshields et~al.(1935)Greenshields, Bibbins, Channing, and
  Miller]{greenshields1935study}
B.~D. Greenshields, J.~Bibbins, W.~Channing, and H.~Miller, ``A study of
  traffic capacity,'' in \emph{Highway research board proceedings}, vol.~14,
  no.~1.\hskip 1em plus 0.5em minus 0.4em\relax Washington, DC, 1935, pp.
  448--477.

\bibitem[Sun et~al.(2014)Sun, Pan, and Gu]{sun2014data}
L.~Sun, Y.~Pan, and W.~Gu, ``Data mining using regularized adaptive b-splines
  regression with penalization for multi-regime traffic stream models,''
  \emph{Journal of Advanced Transportation}, vol.~48, no.~7, pp. 876--890,
  2014.

\bibitem[Amb{\"u}hl et~al.(2017)Amb{\"u}hl, Loder, Menendez, and
  Axhausen]{ambuhl2017empirical}
L.~Amb{\"u}hl, A.~Loder, M.~Menendez, and K.~W. Axhausen, ``Empirical
  macroscopic fundamental diagrams: New insights from loop detector and
  floating car data,'' in \emph{TRB 96th Annual Meeting Compendium of
  Papers}.\hskip 1em plus 0.5em minus 0.4em\relax Transportation Research
  Board, 2017, pp. 17--03\,331.

\bibitem[Liu et~al.(2019)Liu, Liu, Meng, and Cheng]{liu2019tailored}
Z.~Liu, Y.~Liu, Q.~Meng, and Q.~Cheng, ``A tailored machine learning approach
  for urban transport network flow estimation,'' \emph{Transportation Research
  Part C: Emerging Technologies}, vol. 108, pp. 130--150, 2019.

\bibitem[Nam et~al.(2020)Nam, Lavanya, Jayakrishnan, Yang, and
  Jeon]{nam2020deep}
D.~Nam, R.~Lavanya, R.~Jayakrishnan, I.~Yang, and W.~H. Jeon, ``A deep learning
  approach for estimating traffic density using data obtained from connected
  and autonomous probes,'' \emph{Sensors}, vol.~20, no.~17, p. 4824, 2020.

\bibitem[Wang and Papageorgiou(2005)]{wang2005real}
Y.~Wang and M.~Papageorgiou, ``Real-time freeway traffic state estimation based
  on extended kalman filter: a general approach,'' \emph{Transportation
  Research Part B: Methodological}, vol.~39, no.~2, pp. 141--167, 2005.

\bibitem[Seo et~al.(2017)Seo, Bayen, Kusakabe, and Asakura]{seo2017traffic}
T.~Seo, A.~M. Bayen, T.~Kusakabe, and Y.~Asakura, ``Traffic state estimation on
  highway: A comprehensive survey,'' \emph{Annual reviews in control}, vol.~43,
  pp. 128--151, 2017.

\bibitem[Seo et~al.(2015)Seo, Kusakabe, and Asakura]{seo2015estimation}
T.~Seo, T.~Kusakabe, and Y.~Asakura, ``Estimation of flow and density using
  probe vehicles with spacing measurement equipment,'' \emph{Transportation
  Research Part C: Emerging Technologies}, vol.~53, pp. 134--150, 2015.

\bibitem[Xing et~al.(2022)Xing, Wu, Cheng, and Liu]{xing2022traffic}
J.~Xing, W.~Wu, Q.~Cheng, and R.~Liu, ``Traffic state estimation of urban road
  networks by multi-source data fusion: Review and new insights,''
  \emph{Physica A: Statistical Mechanics and its Applications}, vol. 595, p.
  127079, 2022.

\bibitem[Coifman(2002)]{coifman2002estimating}
B.~Coifman, ``Estimating travel times and vehicle trajectories on freeways
  using dual loop detectors,'' \emph{Transportation Research Part A: Policy and
  Practice}, vol.~36, no.~4, pp. 351--364, 2002.

\bibitem[Duan et~al.(2016)Duan, Lv, Liu, and Wang]{duan2016efficient}
Y.~Duan, Y.~Lv, Y.-L. Liu, and F.-Y. Wang, ``An efficient realization of deep
  learning for traffic data imputation,'' \emph{Transportation research part C:
  emerging technologies}, vol.~72, pp. 168--181, 2016.

\bibitem[Di et~al.(2023)Di, Shi, Mo, and Fu]{di2023physics}
X.~Di, R.~Shi, Z.~Mo, and Y.~Fu, ``Physics-informed deep learning for traffic
  state estimation: A survey and the outlook,'' \emph{Algorithms}, vol.~16,
  no.~6, p. 305, 2023.

\bibitem[Huang and Agarwal(2020)]{huang2020physics}
A.~J. Huang and S.~Agarwal, ``Physics informed deep learning for traffic state
  estimation,'' in \emph{2020 IEEE 23rd International Conference on Intelligent
  Transportation Systems (ITSC)}.\hskip 1em plus 0.5em minus 0.4em\relax IEEE,
  2020, pp. 1--6.

\bibitem[Li et~al.(2022)Li, Arora, and Osorio]{li2022fundamental}
Y.~Li, N.~Arora, and C.~Osorio, ``On the fundamental diagram of signal
  controlled urban roads,'' in \emph{Proceedings of the 11th Triennial
  Symposium on Transportation Analysis}, 2022.

\bibitem[LeCun et~al.(2015)LeCun, Bengio, and Hinton]{lecun2015deep}
Y.~LeCun, Y.~Bengio, and G.~Hinton, ``Deep learning,'' \emph{nature}, vol. 521,
  no. 7553, pp. 436--444, 2015.

\bibitem[Clevert et~al.(2015)Clevert, Unterthiner, and
  Hochreiter]{clevert2015fast}
D.-A. Clevert, T.~Unterthiner, and S.~Hochreiter, ``Fast and accurate deep
  network learning by exponential linear units (elus),'' \emph{arXiv preprint
  arXiv:1511.07289}, 2015.

\bibitem[Abadi et~al.(2016)Abadi, Barham, Chen, Chen, Davis, Dean, Devin,
  Ghemawat, Irving, Isard, et~al.]{abadi2016tensorflow}
M.~Abadi, P.~Barham, J.~Chen, Z.~Chen, A.~Davis, J.~Dean, M.~Devin,
  S.~Ghemawat, G.~Irving, M.~Isard \emph{et~al.}, ``$\{$TensorFlow$\}$: a
  system for $\{$Large-Scale$\}$ machine learning,'' in \emph{12th USENIX
  symposium on operating systems design and implementation (OSDI 16)}, 2016,
  pp. 265--283.

\bibitem[of~Public Roads. Office~of Planning. Urban
  Planning~Division(1964)]{us1964traffic}
U.~B. of~Public Roads. Office~of Planning. Urban Planning~Division,
  \emph{Traffic assignment manual for application with a large, high speed
  computer}.\hskip 1em plus 0.5em minus 0.4em\relax US Department of Commerce,
  1964.

\bibitem[Virtanen et~al.(2020)Virtanen, Gommers, Oliphant, Haberland, Reddy,
  Cournapeau, Burovski, Peterson, Weckesser, Bright, et~al.]{virtanen2020scipy}
P.~Virtanen, R.~Gommers, T.~E. Oliphant, M.~Haberland, T.~Reddy, D.~Cournapeau,
  E.~Burovski, P.~Peterson, W.~Weckesser, J.~Bright \emph{et~al.}, ``Scipy 1.0:
  fundamental algorithms for scientific computing in python,'' \emph{Nature
  methods}, vol.~17, no.~3, pp. 261--272, 2020.

\bibitem[Pan and Yang(2009)]{pan2009survey}
S.~J. Pan and Q.~Yang, ``A survey on transfer learning,'' \emph{IEEE
  Transactions on knowledge and data engineering}, vol.~22, no.~10, pp.
  1345--1359, 2009.

\bibitem[Neun et~al.(2023)Neun, Eichenberger, Xin, Fu, Wiedemann, Martin,
  Tomko, Amb{\"u}hl, Hermes, and Kopp]{neun2023metropolitan}
M.~Neun, C.~Eichenberger, Y.~Xin, C.~Fu, N.~Wiedemann, H.~Martin, M.~Tomko,
  L.~Amb{\"u}hl, L.~Hermes, and M.~Kopp, ``Metropolitan segment traffic speeds
  from massive floating car data in 10 cities,'' \emph{IEEE Transactions on
  Intelligent Transportation Systems}, 2023.

\end{thebibliography}

\end{document}